\documentclass[letterpaper]{article} 
\usepackage{aaai22}  
\usepackage{times}  
\usepackage{helvet}  
\usepackage{courier}  
\usepackage[hyphens]{url}  
\usepackage{graphicx} 
\usepackage{natbib}  
\usepackage{caption} 
\DeclareCaptionStyle{ruled}{labelfont=normalfont,labelsep=colon,strut=off} 
\usepackage{hyperref}
\usepackage{algorithm}
\usepackage{algorithmic}
\usepackage{subfigure}
\usepackage{stmaryrd}
\usepackage{multirow}
\usepackage{booktabs}
\usepackage{makecell}
\usepackage{amsmath}
\usepackage{newfloat}
\usepackage{listings}
\usepackage{bibentry}

\floatstyle{ruled}
\newfloat{listing}{tb}{lst}{}
\floatname{listing}{Listing}

\begin{document}

\title{Multi-modal Attention Network for Stock Movements Prediction}
\author{
    Shwai He \textsuperscript{\rm 1}, 
    Shi Gu \textsuperscript{\rm 1}
}
\affiliations{
    \textsuperscript{\rm}University of Electronic Science and Technology of China\\
    shwai.he@gmail.com, gus@uestc.edu.cn\\
}

\maketitle

\begin{abstract}
    Stock prices move as piece-wise trending fluctuation rather than a purely random walk. Traditionally, the prediction of future stock movements is based on the historical trading record. Nowadays, with the development of social media, many active participants in the market choose to publicize their strategies, which provides a window to glimpse over the whole market's attitude towards future movements by extracting the semantics behind social media. However, social media contains conflicting information and cannot replace historical records completely. In this work, we propose a multi-modality attention network to reduce conflicts and integrate semantic and numeric features to predict future stock movements comprehensively. Specifically, we first extract semantic information from social media and estimate their credibility based on posters' identity and public reputation. Then we incorporate the semantic from online posts and numeric features from historical records to make the trading strategy. Experimental results show that our approach outperforms previous methods by a significant margin in both prediction accuracy (61.20\%) and trading profits (9.13\%). It demonstrates that our method improves the performance of stock movements prediction and informs future research on multi-modality fusion towards stock prediction. 
\end{abstract}

\section{Introduction} 
Stock price movements are related to information \cite{Eugene:69}. Traditionally, information affects the movements as news by exposing certain expectations on future trends for the participants in the stock market. Early studies utilized time series of historical trending features to predict price trends \cite{marcek2000stock}. However, stochastic stock prices always hinder predictions. Nowadays, the propensity of social media extremely speeds up the spreading of news and provides an opportunity of estimating the perspective of public attitude towards a certain type of financial asset. For example, the recent dramatic movements of \emph{GameStop} and \emph{Dogecoin} suggest that social media may act as a complementary source to the historical trending in the stock price prediction. 
\par Traditional works mainly take the information like news and exploit the feature engineering to infer its impact on the future price \cite{10.1145/1462198.1462204}. With the prevalence of deep neural networks, stock prediction is studied as an application of NLP downstream tasks, like topic extraction \cite{si-etal-2013-exploiting} and sentiment analysis \cite{nguyen-shirai-2015-topic}. In recent work, researchers applied new learning policies and model structures. \cite{xu-cohen-2018-stock} uses neural variational inference to treat stochasticity in the market better. \cite{feng2019enhancing} enhance the accuracy of prediction through adversarial training policy. \cite{liu-etal-2019-transformer,9159584} resort to attention mechanism and capsule network to mine deeper level of a semantic sequence from social texts. 
\par On the one hand, the social media contexts bring new sources besides the historical data into the prediction model. On the other hand, there also arise some specific challenges, due to the heterogeneity of the social media context. For example, social media texts often contain conflicting and unreliable information \cite{10.1145/3137597.3137600} while analyzing a stock trend requires professional knowledge. Another issue is that conflicting attitudes negatively affect not only consumers' purchasing decisions \cite{vali2015effect} but also models' judgments. Thus, we should resort to multiple resources of data on the text source to evaluate the reliability of provided information \cite{10.1145/3187009.3177739}. The reliability of a social post depends on its poster's reputation on the social media platform and the reader's feedback. Thus we extract social impact features of the two aspects and utilize them to decode posts based on the attention mechanism, which guarantees both semantics and reliability. 
\par Considering the well-established classical financial theories assuming that the instructional information is represented in the price \cite{7065011}, we also resort to the historical trending features reflected in the daily price record. We hypothesize that the historical trending data and social media copra potentially contain complementary information that can only be captured by its modality thus the multi-modal prediction model can improve the predictive model performance. In this article, we propose a novel predictive model of future stock movements by integrating the textual information collected on a financial forum and the historical trending data through the Inter-Intra attention mechanism \cite{peng2019dynamic}, which mimics the combined action of texts and prices. In this way, we use the comprehensive information to infer the whole market perspectives on the related financial asset. We evaluate our model on a stock movements prediction task on our home-collected data and show that our model achieves the SOTA performance. We contribute to build a comprehensive dataset involving time series and texts and propose a novel framework to  utilize reliability estimation and the multi-modality mechanism to predict stock movements. 
\section{Related Work}
Stock trend prediction has already been researched for decades \cite{RePEc:bla:jfinan:v:40:y:1985:i:3:p:793-805}, due to its great value in seeking to maximize stock investment profit. Original approaches are mainly based on time series analysis of historical stock prices. However, because of the excess volatility of stock prices \cite{Timmermann1996ExcessVA}, prediction accuracy is limited. With the development of the Internet and machine learning, financial websites provide large amounts of available textual resources, NLP methods have become the mainstream in this track. \cite{10.1145/3269206.3269290} and \cite{xu-cohen-2018-stock} built corpora from Twitter. \cite{si2013exploiting}. \cite{nguyen-shirai-2015-topic} built topic-based models to predict stock price movements using sentiments on social media by capturing topics and sentiments. \cite{liu-etal-2019-transformer} utilize both transformer and capsule network to deeply mine the textual semantics and infer the movements based on semantics. But these works are hindered by conflicting semantics in the social media and incomplete features of a single modality. 
\par In recent work, some researchers adopted new features and learning policies to overcome the stochastic in the market. \cite{xu-cohen-2018-stock} utilizes both historical trending and social messages to reduce stochastic factors. \cite{feng2019enhancing} proposes an adversarial training policy to add perturbations to simulate the stochasticity of price variable. \cite{9159584} designed a hierarchical network to fuse both financial news and twitters to enrich the information and achieve higher accuracy. Although making a profound process, these methods lack explicit measures to reduce the chaos in social media and fully utilize multi-modality information. 
\par We resort to social impact features and historical trending features to overcome the simplicity of a single modality. Utilizing the attention mechanism, we estimate the credibility of social media posts and incorporate their semantics with objective historical trending features. It improves the results of the stock movements prediction task significantly. 

\section{Problem Formulation}
We aim at predicting the movements of stocks based on the market information at a reference time $t$. The movements lie in a target time window from $t$ to $t + \Delta t$.
The market information we use consists of three parts: social media corpora $C_t$ extracted from financial websites, historical trending data $H_t$ based on daily price, and social impact features $A_t$ that includes the poster features and readers' feedback. The market information lies in a fixed size period $[t-l+1, t]$, and $l$ is the size. Therefore, we formulate the task as follows: 
\begin{equation}
    \hat Y_t=F(C_t, H_t, A_t),
\end{equation}
where $F$ is our model, $\hat Y_t$ is the prediction output. We denote the movements into binary labels $Y_t$ where $1$ represents the rise, $0$ represents the fall, which is as followed: 
\begin{equation}
Y_t=\left\{
\begin{aligned}
1 &  & p_{t+\Delta t} \geq p_t \\
0 &  & p_{t+\Delta t} < p_t  \\
\end{aligned}
\right.,
\end{equation}
where $p_{t+\Delta t}$ denotes the mean adjusted closing price from $d + 1$ to $d + w$, where $w$ is 5 days in our experiment and $p_{t}$ denotes the adjusted closing price in reference time $t$. We observed that there are some targets with exceptionally minor movements ratios due to the uncertainty of the market. In previous stock movements prediction tasks, a common practice is to set upper and lower thresholds on the stock price change and remove samples between them. Similarly, we set two thresholds at $-0.75\%$ and $0.75\%$. 
\section{Data Collection \& Preprocessing}
\textbf{Dataset Description.} Our dataset \footnote{https://github.com/HeathCiff/Multi-modal-Attention-Network-for-Stock-Movements-Prediction} consists of three parts: the social media corpora, the social impact description, and the historical trending data. The social media corpora are extracted from \emph{Xueqiu} \footnote{https://xueqiu.com/}, a popular financial communication platform. Considering the timeliness, we gather the texts released at $t - l + 1$ to $t$ and limit the maximum $n$ number of texts to 96. Since high-trade-volume stocks tend to be discussed more in this forum, we select the top 150 stocks based on their popularity, ensuring relatively sufficient corpora. The historical trending data includes the daily open price, close price, high price, low price, and volume. Inspired by \cite{feng2019enhancing}, we compute and utilize the dispersion between daily high price and low price, between daily open price and close price. In accord with the poster $C_t^i$, the $i$-th text in $C_t$ released at $t_i$, $H_t^i$ is the price trend from $t_i-63$ to $t_i$. Social impact features $A_t$, including the description of posters themselves and readers' feedback, play a role as the supplement of our corpora. The numbers of the poster's fans, followers, and posted texts describe the basic information of posters, while his concerned stocks reflect his correlation with the talked stocks. The poster's transaction performance, such as the profit of invested stocks, reflects his posts' credibility. Readers reply to, like, and retweet the posts, and we record these actions as numeric features by times. We integrate all the social impact features into a vector. Details of the three kinds of features and relevant calculations are shown in Table 1. 
\begin{table}[ht]
	\centering  
	\resizebox{0.45\textwidth}{!}{
	\begin{tabular}{c|c|c}  
		\hline 
		Features&Description&Calculation \\ 
		\hline 
		Social Corpora&texts in financial website& $word2vec$ \\ 
		\hline 
		Historical Trending&historical daily price& chronological arrangement\\ 
		\hline
		\multirowcell{4}{\shortstack{Social Impact Features}}& fans, followers, posted texts, stocks & \multirowcell{2}{\shortstack Concatenation} \\
		& like, retweets and replies & \\
		& stocks of interested & $S_i=Simiarity(S_i,S)$ \\
		& the profit in stocks exchange & 
    $Concat(P_1,P_2,\cdots,P_n)$ \\
		\hline
	\end{tabular}
	}
	\caption{Details of three Features. $S_i$ denotes the name of the $i$-th stock, and $P_i$ denotes the poster's profit of the $i$-th stock. $Similarity$ denotes the similarity between the two word-vectors.}
	\label{table1}
\end{table} 

\textbf{Preprocessing.} For the social media context, we observed that many new words are undetectable by the tokenizer. So we apply left-side and right-side information entropy and mutual information \cite{10.1162/089976603321780272} to extract new words. In some languages, like Chinese, new words generally consist of successive known words. Therefore, we first combine adjacent words and compute the left-side and right-side entropy by summing up the entropy of the word on the left or right side of them: 
\begin{equation}
    H^l_{(x)}=\sum_{y \in left(x)} - p(x) \log p(x), 
\end{equation}
\begin{equation}
    H^r_{(x)}=\sum_{y \in right(x)} - p(x) \log p(x), 
\end{equation}
where $H^l_{(x)}$ and $H^r_{(x)}$ denote left-side and right-side information entropy, and $p$ denotes the probability of occurrence in a sentence. $x$ is a potential new word, and $y$ is the word in the left (or right) of $x$. Comparing with a threshold, we keep words that have higher information entropy values. Then we compute the mutual information of internal parts in kept words: 
\begin{equation}
    I_{(x)}=\sum_{a, b \in x} p(a, b) \log \frac{p(a, b)}{p(a) p(b)},
\end{equation}
where $a$ and $b$ are the two parts of a word. We select top-k words based on mutual information values and add them to our dictionary. 
\par The lengths of texts are usually uneven. Long texts exceed our maximum length restriction, while short ones may be meaningless. Therefore, we remove short texts and put the remains into key sentences extraction processing \cite{INR-019} to drop off meaningless sentences and make their lengths fit for our model. In extraction processing, we first filter out the stop words and then apply a \emph{BM25} algorithm \cite{10.1561/1500000019} to estimate the similarity between the whole text and every single sentence within it and remain sentences with the high \emph{BM25} values. 
\section{Our Model}
\subsection{Model Overview}
The architecture of our framework is illustrated in Figure 1, which consists of four sequential modules:
\begin{enumerate}
\item \textbf{Embedding Module} embeds textual features, historical trending features, and social impact features into vectors. 
\item \textbf{Encoder Module} encodes textual features to acquire high-level semantics. 
\item \textbf{Fusion Module} complements social impact features to textual features and mix up the fused features with historical trending features into an intact feature map.  
\item \textbf{Inference Module} predicts the stock movements based on the intact feature map. 
\end{enumerate}
\begin{figure*}[htbp]
    \centering
    \includegraphics[width=0.95\textwidth]{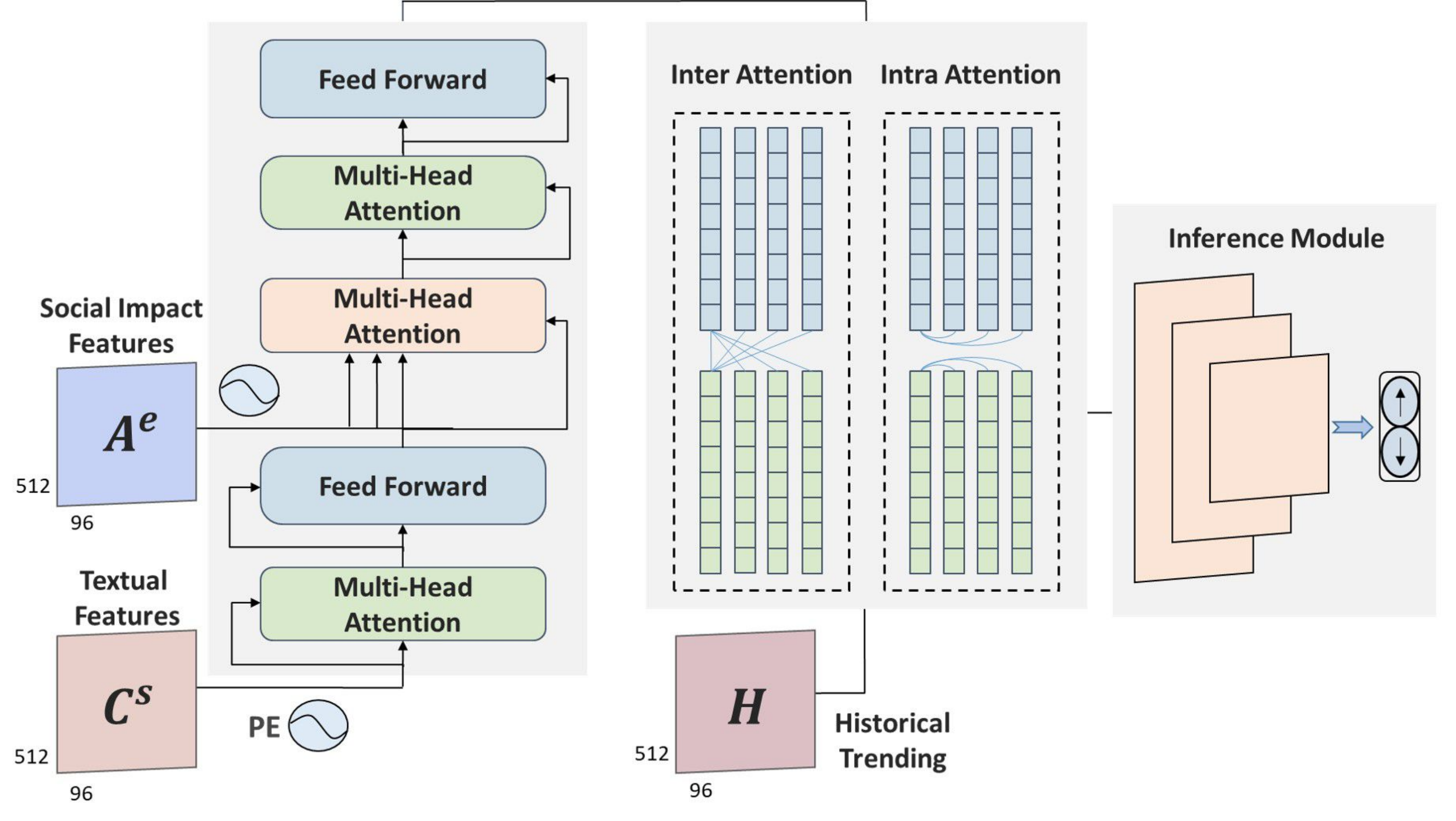}
    \caption{Illustration of our network architecture. The lower left is the Encoder Module. We learn the semantics of textual features $C^s$ using
a transformer encoder. The upper left illustrates the fusion process of textual features $C^e$ and social impact features $A^e$. The meddle part denotes the fusion of historical trending $H$ and decoded textual features $C$. In the right part Inference Module, our model predicts future stock movements, rising or falling.}
\end{figure*}
\subsection{Embedding Module}
Embedding Module embeds three modalities. As to each social media post $C_t$, there is a relevant historical trending feature $H_t$ and a social impact feature $A_t$. Therefore, we define the temporal input $X_t$ as follows: 
\begin{equation}
    X_{t}=\left[C_{t}, H_{t}, A_{t}\right]. 
\end{equation} 
To represent the semantics of a text, we feed $C_t \in R^{n \times s \times d}$ into a deep averaging network \cite{iyyer-etal-2015-deep} to acquire text-level vectors $C^s \in R^{n \times d}$, where $n$ denotes the length of a sequence, $s$ denotes the fixed text length in words, $d$ denotes the latent dimension of every word vector. In our experiment, we find the sum-up operation may lead to an information loss. So we replace it with a flatten operation to maintain affluent information. As for the social impact features, we project it into high-dimension features $A^e \in R^{n \times d}$, which maintains adequate information and is compatible with the fusion with textual features. 
\begin{equation}
    C^{s}=DAN(C_{t}), 
\end{equation}
\begin{equation}
    A^{e}={\rm Linear}(A_{t}), 
\end{equation}
where $DAN$ denotes the deep averaging network, \text{Linear} denotes fully connected layers. Each historical trending feature is denoted as a three-dimensional feature map $H_t \in R^{n \times 64 \times 7}$, where 64 is the length of the historical price range in days, and $7$ is equal to the number of daily price features like open price. We resort to 3D-CNNpred \cite{HOSEINZADE2019273} to fully extract the spatial information $H \in R^{n \times d}$: 
 \begin{equation}
    H=CNNpred(H_{t}). 
\end{equation} 
\subsection{Encoder Module}
Transformer encoder has been used widely in exploring textual semantics, and we adopt it in this module. Inspired by the multi-head attention mechanism \cite{vaswani2017attention}, we utilize it to represent the mutual connection of each part of the time-based textual sequence and positional-wise feed it forwards to the next layer. In this way, we capture social attitudes at the textual level: 
\begin{equation}
C^{e}=TransEncoder(C^s_q, C^s_k, C^s_v),
\end{equation}
where $TransEncoder$ denotes the encoder in the transformer, $C^e$ denotes the encoded textual features. $C^s_q$, $C^s_k$, $C^s_v$ identifies the queries, keys, values of $C^s$, respectively. 
\par In recent works, the time feature was important because of the timeliness of social posts. To fully utilize this feature, we design a time-based positional encoding method to maintain temporal information. Like positional encoding in the transformer \cite{DBLP:journals/corr/QinSCCJC17}, we replace the relative position with the relative dispersion between posts' release time and the reference time: 
\begin{equation}
    P E_{(t, 2 i)} =\sin 
    (\frac{t}{c^{2 i / d}}), 
\end{equation} 
\begin{equation}
    P E_{(t, 2 i+1)} =\cos 
    (\frac{t}{c^{2 i / d}}), 
\end{equation}
where $PE$ denotes the positional encoding, $t$ denotes the relative time dispersion, $i$ denotes the $i$-th position in the vector. $c$ is a free parameter, and we set it as $10000$, which is the same with \cite{vaswani2017attention}. $d$ is the vectors' latent dimension. 
\subsection{Fusion Module}
\textbf{Text and Social Impact Features.} Exploring the textual semantics is not enough, for the reliability varies. On the one hand, posters having many fans and followers tend to perform well in this forum. It is related to the high reliability of their attitudes on stock movements. On the other hand, in common sense, well-received texts are valid and truthful. 
\par In the previous work, \cite{liu-etal-2019-transformer} utilized the transformer encoder, which distinguishes the task-related importance through measuring the similarity of each part. However, resting on texts themselves is limited, so we resort to social impact features to replenish and apply an attention mechanism-based decoder to redistribute the importance of the textual sequence: 
\begin{equation}
C = TransDecoder(A^e_q, A^e_k, C^e_v), 
\end{equation} 
where $TransDecoder$ denotes the decoder in the transformer, $C$ denotes the decoded textual features, and $C^e_v$ denotes the values of $C^e$. $A^e_q$, $A^e_k$ identifies queries and keys of $A^e$, respectively. 
\par The decoder architecture is consistent with the decoder in the transformer \cite{vaswani2017attention}. In encoder-decoder multi-head attention, we project Encoder Module output $C^e$ into values, project social impact features $A^e$ into keys and queries. We perform the attention function on each combination that consists of queries, keys, and values in parallel, yielding output values. In this way, we treat textual features with different social impact features separately so that the fused outputs $C$ maintain semantics and reliability. 
\par \textbf{Historical Trending and Fused Text.}
Generally, before investing, investors usually refer to historical trending and social media to get a comprehensive guide, which means they are not equal totally but complementary. Therefore, we hypothesize that historical trending data supplement the social media texts and vice versa. Instead of concatenating them simply, we describe this supplement effect as an information flow and update the two modalities through the inter-attention mechanism. Because of the Markov property in historical price \cite{7065011} and frequent interaction on the Internet, each feature is not independent of others within the same modality, so there also exists an internal information flow. Therefore, we adopt an intra-mechanism to represent the information flow within the updated modalities. 
\par The inter attention block computes the bidirectional supplement between historical trending and textual information. Like Encoder Module, we first transform each historical trending feature and textual feature into queries, keys, and values. Then we compute the transfer values between the two modalities: 
\begin{equation}
H^{itv}=\operatorname{softmax}(\frac{H_q \times C_k^{T}}{\sqrt{d}}) \times C_v, 
\end{equation}
\begin{equation}
C^{itv}=\operatorname{softmax}(\frac{C_q \times H_k^{T}}{\sqrt{d}}) \times H_v. 
\end{equation} where we denote the transformational historical trending features as $H_k, H_q, H_v \in R^{n \times d}$ and transformational textual feature as $C_k, C_q, C_v \in R^{n \times d}$. $H^{itv}$ and $C^{itv}$ denote the transfer values in inter-attention block. $T$ denotes the transpose operation. $\times$ denotes the matrix multiplication, $d$ denotes the dimension of a feature vector. We update two modalities by projecting the concatenation of original values and transfer values into $d$-dimensional values: 
\begin{equation}
H^{itd}=\operatorname{Linear}(\left[H, H^{itv}\right]^{T}), 
\end{equation}
\begin{equation}
C^{itd}=\operatorname{Linear}(\left[C, C^{itv}\right]^{T}). 
\end{equation}
where $H^{itd}$ and $C^{itd}$ are the updated values of the two modalities in this block, respectively. We also project $H^{itd}$ and $C^{itd}$ into queries, keys, and values. To summarize the conditioning information from the other modality, we average-pool the two modalities
along the sequence-index dimension and project them to match the dimension of the queries and keys. Then we feed them to a sigmoid non-linearity function to generate channel-wise conditioning gates for the other modality. We utilize the conditioning gates to update original queries and keys: 
\begin{equation}
G_{H}=\sigma (\text{Linear} (\operatorname{Avg\_Pool}(C^{itd}))), 
\end{equation}
\begin{equation}
\hat{H}_q^{itd}=(1+G_{H}) \odot H_q^{itd}, \hat{H}_k^{itd}=(1+G_{H}) \odot H_k^{itd}, 
\end{equation}
\begin{equation}
G_{C}=\sigma (\text{Linear} (\operatorname{Avg\_Pool}(H^{itd}))), 
\end{equation}
\begin{equation}
\hat{C}_q^{itd}=(1+G_{C}) \odot C_q^{itd}, \hat{C}_k^{itd}=(1+G_{C}) \odot C_k^{itd}. 
\end{equation}
where $\sigma$ denotes sigmoid function, $\odot$ denotes element-wise multiplication, $G$ denotes the output of sigmoid door gating, $\hat{H}_q^{itd}$ and $\hat{H}_k^{itd}$ denotes the updated queries and keys of historical trending modality, $\hat{C}_q^{itd}$ and $\hat{C}_k^{itd}$ denotes the updated queries and keys of textual modality. 
\par The intra-attention block captures the comprehensive understanding of each updated modality, considering the internal influence. We utilize an attention mechanism to update the values based on updated keys and queries like the inter-attention block. Then we feed the summation of updated values ($H^{inu}$ and  $C^{inu}$) and original values ($H^{itd}$ and $C^{itd}$) into a fully connected layer: 
\begin{equation}
H^{inu} =\operatorname{softmax}(\frac{\hat{H}_q^{itd} \times
\hat{H}_k^{{itd}^T}}{\sqrt{d}}) \times H_v^{itd}, 
\end{equation}
\begin{equation}
H^{ind} = \operatorname{Linear} (H^{itd} + H^{inu}), 
\end{equation}
\begin{equation}
C^{inu} =\operatorname{softmax}(\frac{\hat{C}_q^{itd} \times
\hat{C}_k^{{itd}^T}}{\sqrt{d}}) \times C_v^{itd}, 
\end{equation}
\begin{equation}
C^{ind} = \operatorname{Linear} (C^{itd} + C^{inu}). 
\end{equation}
where $H^{ind} \in R^{n \times m \times h}$ and $C^{ind} \in R^{n \times m \times h}$ are the output of the intra-attention block, for we maintain the multi-head form. Then we average pool the element-wise product of $C^{ind}$ and $H^{ind}$ along the sequence-index dimension into the comprehensive feature map $FM$: 
\begin{equation}
    FM = \operatorname{Avg\_Pool}(H^{ind} \odot C^{ind}). 
\end{equation}
\subsection{Inference Module}
With all three modalities fused, we obtain a sequence of comprehensive feature map $FM \in R^{m \times h}$ in chronological order, where $m$ denotes the number of heads and $h$ denotes the length of each head. We design 1D convolution layers to make the final inference, down-sampling the feature map by several kernels and pooling it to generate the prediction arise from the comprehensive information: 
\begin{equation}
    \hat Y=Conv(FM), 
\end{equation}
where $\hat Y$ denotes the output and $Conv$ denotes the convolution layers. We choose the marginal loss function \cite{sabour2017dynamic} on the prediction and the target label, and the loss is propagated back from the convolution layers' outputs across the entire model. 
\begin{equation}
\begin{array}{l}
 { Loss_k }=Y_k \max (0, m^{+}-\|\hat{Y}_k\|)^{2} \\
    +\lambda\left(1-Y_k\right) \max (0,\|\hat{Y}_k\|-m^{-})^{2}, 
\end{array}
\end{equation}
where $Loss_k$ denotes the loss of the $k$-th category,$\lambda$ is $0.5$, $m^{+}$ is $0.9$ and $m^{-}$ is $0.1$. $\hat{Y}_k$ is the probability of the $k$-th category of the prediction while ${Y}_k$ is $k$-th element of the label. 
\par Inspired by \cite{sabour2017dynamic}, we take reconstruction as a regularization method. The prediction in this part is fed into fully connected layers to reconstruct the input of the Inference Module. By minimizing the sum of squared differences between them, the Inference Module can be robust. 
\begin{equation}
    Loss=\sum Loss_k+\lambda (FM-Re(\hat{Y}))^2, 
\end{equation}
where the first term is the margin loss, and the second is the reconstruction loss, $Re$ is the reconstruction function acting on the output $\hat Y$. $\lambda$ is the scale factor, which is $0.0005$ in our experiment. We scale down the reconstruction loss so that it does not dominate the margin loss during training. 
\section{Experiment}
\subsection{Training Setup}
We restrict 14 days for a textual sample and set the duration of the historical trending description as 64-day. The prediction interval is five days, and there are $64$ shuffled samples in a batch. The maximal length of the text is $64$ in words. And the maximal length of a textual sequence $s$ is restricted to $96$, with excess clipped. To improve the expression of word vectors, we set the latent dimension as $512$. All weight matrices in the model are initialized with the fan-in trick, while we set biases as zero in the beginning. We train the model as an Adam optimizer with an initial learning rate of $0.001$ and follow a linear decay strategy. We use the input dropout rate of $0.2$ and the weight decay rate of $0.001$ to regularize. 
\subsection{Evaluation Metrics}
Following previous work for stock prediction \cite{xie-etal-2013-semantic}\cite{xu-cohen-2018-stock}, we adopt the standard measure of accuracy and Matthews Correlation Coefficient (MCC) as evaluation metrics. With the confusion matrix which contains the number of samples classified as true-positive $tp$,  false-positive $fp$, true-negative $tn$ and false-negative $fn$, MCC is calculated as follows:
\begin{small}
\begin{equation}
MCC=\frac{t p \times t n-f p \times f n}{\sqrt{(t p+f p)(t p+f n)(t n+f p)(t n+f n)}}.
\end{equation}
\end{small}
\par We list the accuracy and MCC performance of baselines and our models in Table 2.
\subsection{Baselines}
To evaluate the effectiveness of our proposed deep learning framework, we compare our model MMAN(Multi-Modality Attention Network) with the performance of the following methods:
\begin{itemize}
\item[$\bullet$]Random Forest: \cite{pagolu2016sentiment} proposes a discriminative Random Forest classifier using word2vec text representations. 
\item[$\bullet$]HAN: \cite{Hu_2018} proposes a state-of-the-art discriminative deep neural network with hierarchical attention. 
\item[$\bullet$]Adv-LSTM: \cite{feng2019enhancing} proposes the attentive LSTM model and train it in adversarial training strategy.  
\item[$\bullet$]StockNet: \cite{xu-cohen-2018-stock} proposes a deep generative model which jointly exploiting text and price signals. 
\item[$\bullet$]MHACN: \cite{9159584} first split corpus into news and tweets and then jointly exploit their combined effect. To keep consistent with our dataset, we take textual features and historical trending features as inputs. 
\item[$\bullet$]CapTE: \cite{liu-etal-2019-transformer} proposes a CapTE model to consider the rich semantics and relation for a certain stock. 
\item[$\bullet$]MMAN-oC: We only take textual features as the input to compare the modality fusion mechanism. 
\item[$\bullet$]MMAN-oH: We only take historical trending features as the input. 
\item[$\bullet$]MMAN-nA: To verify the effectiveness of the social impact features, we conduct experiments by utilizing our model without this modality. 
\item[$\bullet$]MMAN-nH: To verify the effectiveness of the historical trending features, we conduct experiments by utilizing our model without this modality.
\end{itemize}
\subsection{Experimental Results}
In this section, we detail our experimental setup and results. We test our model from two aspects. One is to predict the rise and fall based on our dataset. The other is virtual trading which mimics real-world transactions. In the end, we analyze the effects of multi-modality. 
\par \textbf{Classification Task.} From table 2, we can see that CapTE gets the highest score in the baselines. However, on the same dataset, our model obtains higher accuracy than CapTE, more than $1.33\%$, which shows our model outperforms baselines by a significant margin in both accuracy and MCC. We have provided a detailed discussion of the necessity of access the reliability of texts and our approach to supplement semantics with reliability. The score of MMAN is higher than MMAN-nA, and MMAN-nH is higher than MMAN-oC. It indicates the supplementary effect from social impact features to textual features, which reduces the conflict of a single text modality. As introduced above, we hypothesize that historical trending features and textual features are complementary. In the ablation study, we compare their combined action with an individual effect. We found that our model MMAN outperforms MMAN-nH and MMAN-nA outperforms both MMAN-oC and MMAN-oH, which indicates the two modalities supplement each other and the fused features are more predictive. Compared with each partial model mentioned above, intact model MMAN performs much better, which means our hypothesis is solid and the effectiveness of multi-modality. 
\begin{table}[ht]
\centering
\begin{tabular}{ccc}
\hline Model & Acc. & MCC \\
\hline RF & $53.13 \%$ & $0.0129$ \\
HAN & $55.96 \%$ & $0.0447$ \\
StockNet & $57.35 \%$ & $0.0621$ \\
Adv-LSTM & $57.93 \%$ & $0.0672$ \\
MHACN & $58.84 \%$ & $0.0721$ \\
CapTE & $59.87 \%$ & $0.0976$ \\
Ours-MMAN-oH & $56.67 \%$ & $0.0583$ \\
Ours-MMAN-oC & $59.49 \%$ & $0.0737$ \\
Ours-MMAN-nA & $60.06 \%$ & $0.0850$ \\
Ours-MMAN-nH & $60.46 \%$ & $0.0937$ \\
Ours-MMAN & $\mathbf{61.20 \%}$ & $\mathbf{0.1193}$ \\
\hline
\end{tabular}
\caption{Performance of baselines and MMAN variations in accuracy
and MCC.}
\end{table} 
\par \textbf{Virtual Trading.} In addition to comparing the accuracy and MCC, we apply our model and CapTE to virtual trading and compare their performance. We simulated real-world stock trading from January 2021 to March 2021 by following the strategy proposed by \cite{Lavrenko00miningof}, which mimics the behavior of a trader who uses our model. On the one hand, if the model indicates that an individual stock price will increase, the fictitious trader will invest in \$10,000 worth at the opening price. Then the trader will hold the stock during the prediction internal. During this time, if the stock has a profit of 2\% or more, the trader sells immediately. Otherwise, the trader sells the stock at the closing price on the last day. On the other hand, if the model predicts that a stock price will fall, 
when the trader can buy the stock at a price 1\% lower than shorted, he/she will buy it to cover. Otherwise, the trader buys the stock at the closing price of the last day. 
\begin{table}[ht]
\centering
\begin{tabular}{llll}
\hline
\textbf{Industry} & \textbf{Market} & \textbf{CapTE} & \textbf{MMAN}\\
\hline
Medicine  & \$ 102.6 & \$ 959.2 & \$ 866.6\\
Commerce  & \$ 956.2 & \$ 1051.9 & \$ 1240.8\\
Catering  & \$ 527.0 & \$ 654.7 & \$ 778.8\\ 
Electronic  & \$ 1003.5 & \$ 752.2 & \$ 913.2\\ 
Chemical  & \$ 689.9 & \$ 833.0 & \$ 958.3\\
Transportation & \$ 576.5 & \$ 702.6 & \$ 723.1\\
\hline
\end{tabular}
\caption{\label{citation-guide}
 Profit comparison between \textbf{CapTE} and \textbf{MMAN}. We also consider the close price change between the start and end of the test time window, which is denoted as \textbf{Market}.}
\end{table} 
We split the stocks into categories based on their industries and show the virtual trading performance of six selected popular industries in table 3. We found that the profits of CapTE and our model MMAN are much higher than the price change during the test period. Compared to CapTE, our model performs much better overall, and the maximum return of stocks in the commerce industry is over 12\%. The results demonstrate consistently better performance, which indicates the robustness of our model. 
\subsection{Effects of Multi-Modality}
We utilize the class activation map \cite{zhou2015learning} to visualize the feature maps of outputs of the Encoder Module and Fusion Module, which is illustrated in Figure 2. 
\begin{figure}[ht]
\centering
\subfigure[]{
\begin{minipage}[t]{\linewidth}
\centering
    \includegraphics[scale=0.28]{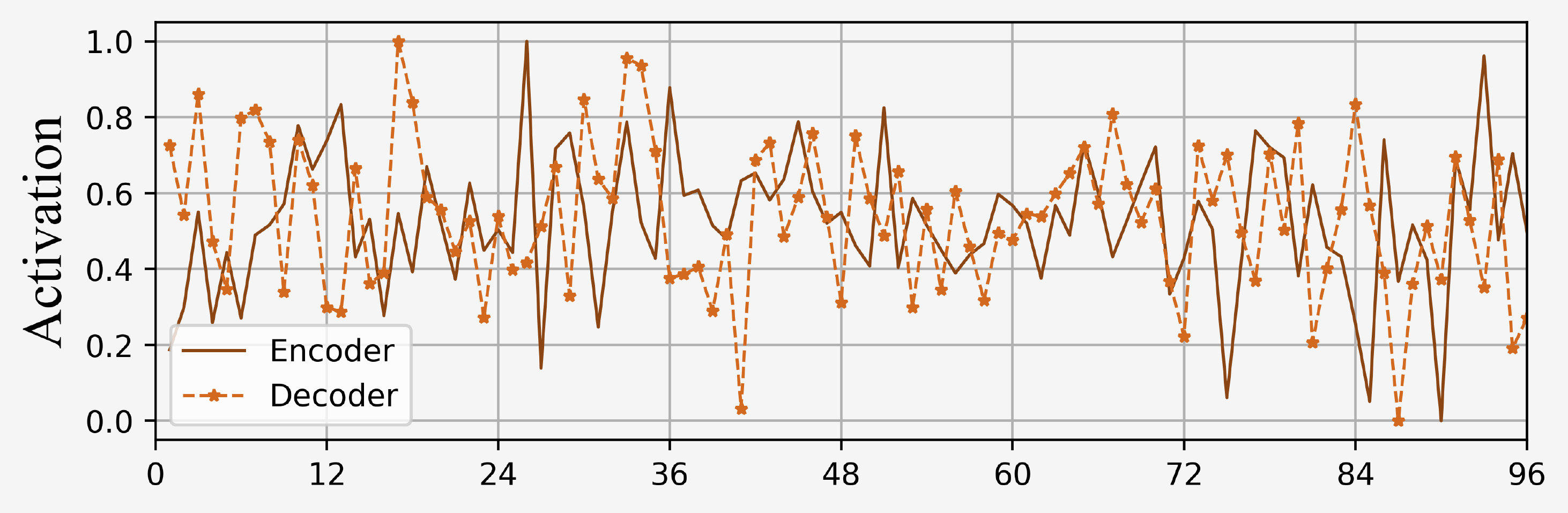}\hfill
\end{minipage}}
\subfigure[]{
\begin{minipage}[t]{\linewidth}
\centering
    \includegraphics[scale=0.28]{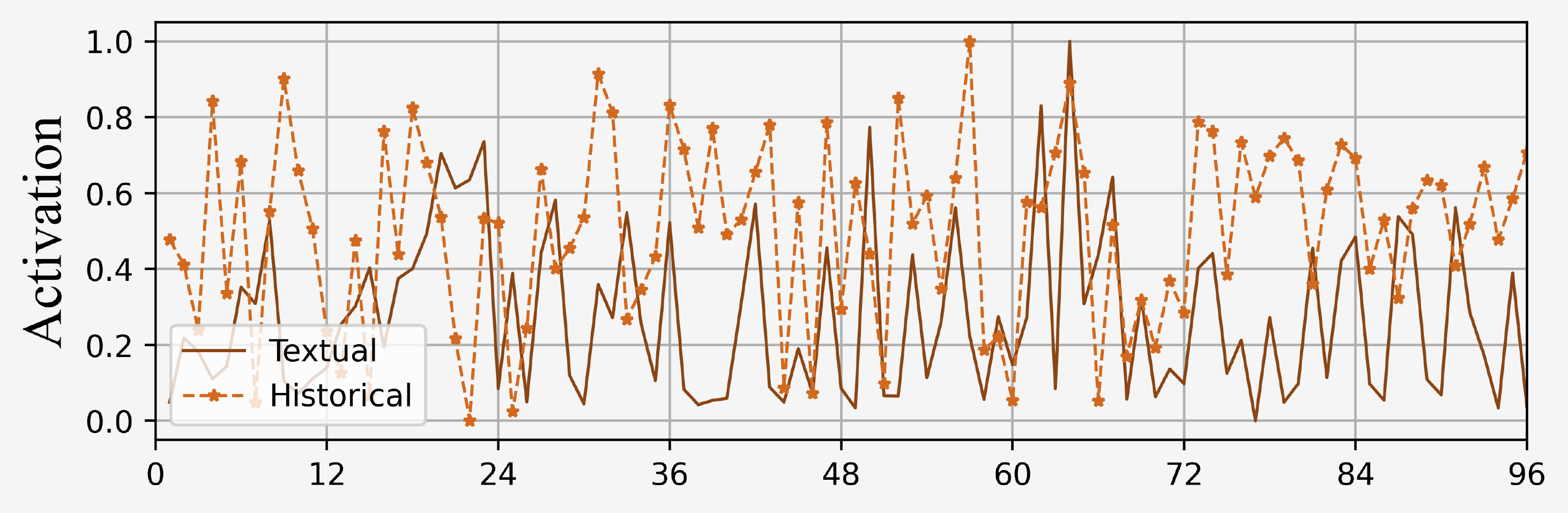}
\end{minipage}}
\caption{Given the same input into our model, the acquired CAM images are as following: (a), the outputs of the Encoder and Decoder, $C^e$ and $C$; (b), the textual modality output $C^{ind}$ and the historical trending modality output $H^{ind}$ of inter-intra attention blocks. The abscissa represents the feature sequence, where $i$ denotes the $i$-th vector in the feature map. The vertical coordinate represents the activation intensity.}
\end{figure} 
\par In the feature maps, we paint every feature vector with gray values based on its activation intensity, where the mean gray value of a one-dimensional feature vector represents the whole vector. Comparing $C^e$ and $C$, we observed that high-intensity parts shifted, which means the fused textual features $C$ do not rest on semantics itself but take in the social impact features. We also compare the textual features $C^{ind}$ and historical trending $H^{ind}$ passed through the inter-intra attention mechanism. We found that $C^{ind}$ and $H^{ind}$ focus on different areas, which is consistent with our hypothesis that historical trending is not equal but supplement to the social attitude in this task. 
\section{Conclusion}
To promote prediction accuracy is to make full use of stock market information. We demonstrate a method to predict the stock price movements based on enriched market information through the extraction and fusion of multi modalities. As is shown in the results, our Multi-Modality Attention Network improves the performance in both the classification task and virtual trading significantly, which introduces a new direction to increase the performance: adopt multi-modality. Since the market is complex, digging for sufficient modalities can better represent the real-world situation. Our work also highlights a question on how to perform an effective fusion among modalities to improve performance. In the future, we will further explore relevant features and high-efficient modality fusion approaches. 
\bibliography{Main}
\end{document}